\pdfoutput=1
\documentclass[conference]{IEEEtran}

\usepackage{cite}
\usepackage{amsmath,amssymb,amsfonts}
\usepackage{algorithmic}
\usepackage{graphicx}
\usepackage{textcomp}
\usepackage{xcolor}
\def\BibTeX{{\rm B\kern-.05em{\sc i\kern-.025em b}\kern-.08em
		T\kern-.1667em\lower.7ex\hbox{E}\kern-.125emX}}

\usepackage{url}
\usepackage[utf8]{inputenc}
\usepackage[small]{caption}
\usepackage{booktabs}
\usepackage{algorithm}
\usepackage{algorithmic}
\usepackage{mathtools}
\usepackage{balance}
\usepackage[OT1]{fontenc}
\usepackage{pifont}
\usepackage{caption}
\usepackage{graphicx}
\usepackage{subcaption}
\usepackage{amsmath}

\usepackage{booktabs}
\usepackage{multirow}
\usepackage{array}

\newcommand{\tabincell}[2]{\begin{tabular}{@{}#1@{}}#2\end{tabular}} 
\usepackage{threeparttable}
\usepackage{multicol}
\usepackage{numprint}
\npthousandsep{\,}

\usepackage{verbatim}
\usepackage{pifont}
\usepackage{xcolor}
\definecolor{grey}{rgb}{0.9,0.9,0.9}
\definecolor{lightgreen}{HTML}{bae4b3}
\definecolor{lightgrey}{HTML}{f0f0f0}
\definecolor{mygreen}{HTML}{31a354}
\definecolor{mygray}{HTML}{666666}

\mathchardef\mhyphen="2D

\newcommand*{\eg}{e.g., }

\newcommand*{\ie}{i.e.,}

\newcommand*{\all}{et al. }

\title{Testing Deep Learning Models for Image Analysis Using Object-Relevant Metamorphic Relations}

\author{\IEEEauthorblockN{Yongqiang Tian\IEEEauthorrefmark{1}, Shiqing Ma\IEEEauthorrefmark{2}, Ming Wen\IEEEauthorrefmark{1}, Yepang Liu\IEEEauthorrefmark{3}, Shing-Chi Cheung\IEEEauthorrefmark{1} and Xiangyu Zhang\IEEEauthorrefmark{4}}
\IEEEauthorblockA{\IEEEauthorrefmark{1}
	Department of Computer Science and Technology\\
	Hong Kong University of Science and Technology, Hong Kong, China\\
	Email: \{ytianas, mwenaa, scc\}@cse.ust.hk}
\IEEEauthorblockA{\IEEEauthorrefmark{2}
	Department of Computer Science\\
	Rutgers University, New Jersey, USA\\
Email: shiqing.ma@rutgers.edu}
\IEEEauthorblockA{\IEEEauthorrefmark{3}
	Shenzhen Key Laboratory of Computational Intelligence\\
	Southern University of Science and Technology, Shenzhen, China\\
Email: liuyp1@sustech.edu.cn}
\IEEEauthorblockA{\IEEEauthorrefmark{4}
	Department of Computer Science\\
	Purdue University, Indiana, USA\\
	Email: xyzhang@cs.purdue.edu}}
\begin{document}
	\maketitle
	\begin{abstract}
Deep learning models are widely used for image analysis.
While they offer high performance in terms of accuracy, people are concerned about if these models inappropriately make inferences using irrelevant features that are not encoded from the target object in a given image.
To address the concern, we propose a metamorphic testing approach that assesses if a given inference is made based on irrelevant features. 
Specifically, we propose two novel metamorphic relations to detect such inappropriate inferences.
We applied our approach to 10 image classification models and 10 object detection models, with three large datasets,~\ie~ImageNet, COCO, and Pascal VOC.
Over 5.3\% of the top-5 correct predictions made by the image classification models are subject to inappropriate inferences using irrelevant features. 
The corresponding rate for the object detection models is over 8.5\%. 
Based on the findings, we further designed a new image generation strategy that can effectively attack existing models.
Comparing with a baseline approach, our strategy can double the success rate of attacks. 
\footnote{Please note that a later version of this paper is accepted by Empirical Software Engineering in 2021.
The title of the accepted paper is \textit{To What Extent Do DNN-based Image Classification Models Make Unreliable Inferences?}.
Please contact the first author if you are interested in the accepted version.
}
\end{abstract}

\section{Introduction}
\label{sec:intro}
Deep learning models have been widely deployed for image analysis applications, such as image classification~\cite{resnet, vgg,mobilenet}, object detection~\cite{ssd, fasterrcnn, yolov3} and human keypoint detection~\cite{DBLP:conf/eccv/XiaoWW18, DBLP:conf/cvpr/ChenWPZYS18, DBLP:journals/corr/abs-1904-08900,peng2018megdet}.
While these image analysis models outperform classical machine learning algorithms, recent studies~\cite{lime, DBLP:journals/corr/abs-1711-11443, DBLP:journals/corr/Moosavi-Dezfooli15} have raised concerns on such models' reliability. 

\begin{figure}[b!]
	\vspace{-3mm}
	\centering
	\begin{subfigure}[b]{0.15\textwidth}
		\includegraphics[width=\textwidth]{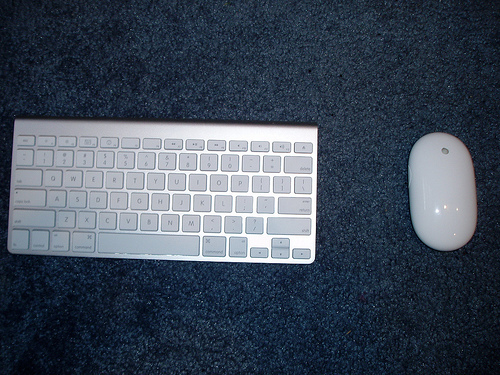}
		\caption{}
		\label{fig:mouse_intro}
	\end{subfigure}
	\begin{subfigure}[b]{0.15\textwidth}
		\includegraphics[width=\textwidth]{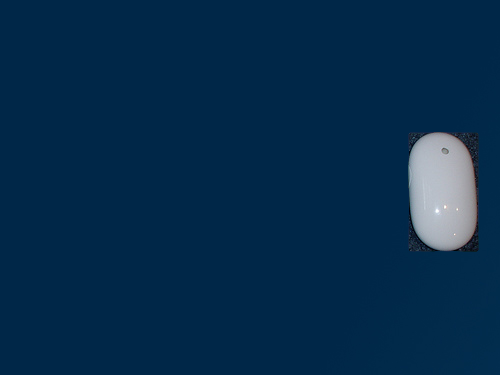}
		\caption{}
		\label{fig:mouse_intro2}
	\end{subfigure}
	\begin{subfigure}[b]{0.15\textwidth}
		\includegraphics[width=\textwidth]{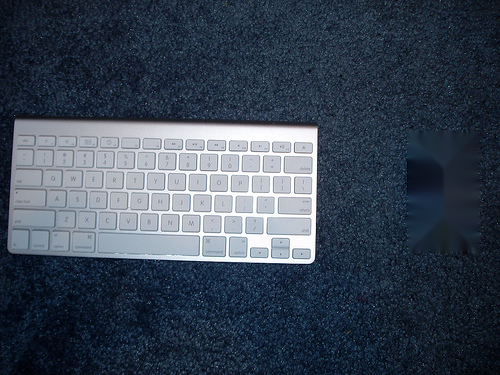}
		\caption{}
		\label{fig:mouse_intro3}
	\end{subfigure}
	\caption{(a): An Image from ImageNet. (b): Object (Mouse) Preserving Mutation. (c): Object (Mouse) Removing Mutation}
\end{figure}

Various testing techniques~\cite{deepxplore, deeptest, Zhang:2018:DGM:3238147.3238187, XIE2011544, issta18meta, 7961649, deepmutation} have been proposed to help assess the reliability of deep learning models for image analysis.
For instance, Pei~\all~\cite{deepxplore} proposed an optimization strategy to generate test inputs for image classification and digit recognition applications.
However, a major limitation of these techniques is that they do not consider whether the inferences made by a model are based on the features encoded from the target objects instead of those encoded from these objects' background. 
We refer to the former as \textit{object-relevant features} and the latter as \textit{object-irrelevant features}. 
For example, those features encoded from the rectangular region occupied by the keyboard object in the image as shown in~\figurename{\ref{fig:mouse_intro}} is considered as object-relevant for a keyboard detection model. 
Other features encoded from the rest of this image are object-irrelevant.
Such relevant and irrelevant features vary with target objects.
For example, a mouse detection model would consider those features encoded from the ``mouse'' in ~\figurename{\ref{fig:mouse_intro}} as object-relevant. 

Deep learning models do not necessarily make inferences based on object-relevant features. 
For instance, a recent study showed that a model would classify an image with bright background as ``wolf'' regardless of the objects in the image~\cite{lime}. 
Even though the model could output accurate results with respect to the test inputs (\ie~test images), such inferences are unreliable.
More seriously, unreliable inferences based on object-irrelevant features are vulnerable to malicious attacks. 
For example, Gu~\all~\cite{badnet} showed that attackers could inject a backdoor trigger, such as a yellow square in an image's background, to a deep neural network (DNN) model.
A model that makes inferences based on object-irrelevant features (e.g., yellow square at the background), will then classify an image containing this trigger to a specific label, regardless of the target object in the image.
Deploying such a model in mission-critical applications could cause catastrophic consequences. 
Therefore, it is important to develop effective techniques to assess deep learning inference results from the perspective of object relevancy.

However, there are two major challenges that prevent us from easily validating inference results generated by deep learning models with respect to object relevancy.
First, obtaining oracles for testing deep learning models is hard \cite{XIE2011544}.
We resort to metamorphic testing~\cite{chen1998metamorphic} to tackle this challenge. 
Metamorphic testing has been popularly leveraged to test deep learning models for image analysis~\cite{Zhang:2018:DGM:3238147.3238187, XIE2011544, issta18meta, 7961649}.
Second, it is difficult to measure whether a model makes inference based on object-relevant features.
Efforts have been made to explain whether an inference made by a model is trustable \cite{lime}.
However, model explanation is still an outstanding challenge.
Recent research focuses mostly on image classification models. 
Further, existing studies cannot quantitatively measure to what extent an inference is made with respect to object relevancy.
Besides, they are designed for the purpose of interpretation instead of testing, and thus cannot be easily adapted to validate inferences made by deep learning models (\eg cannot generate test inputs).
To address this challenge, we propose two novel metamorphic relations (MRs) to quantitatively assess a model's inferences from the perspective of object relevancy as follows:
\begin{itemize}
\item \textbf{MR-1}: An image after altering the regions unoccupied by the target object should lead to a similar inference result. 
\item \textbf{MR-2}: An image after removing the target object should lead to a dissimilar inference result.
\end{itemize}

Formulation of these two metamorphic relations is given in Section~\ref{sec:relation}.
Based on these two relations, we propose a metamorphic testing technique to assess whether an inference made by a deep learning model for image analysis is based on object-relevant features. 
Essentially, we design image mutation operations concerning the two relations to generate test inputs. 
To test the object relevancy of an image inference, we apply these operations on the given image to construct mutated images and check whether the subsequent inferences on such mutated images satisfy the metamorphic relations.
Based on the metamorphic testing results, we devise an ``object-relevancy score'' as a metric to measure the extent to which an inference made by a deep learning model is based on object-relevant features.

We evaluated our technique using 10 common image classification and 10 object detection models on 3 popular large datasets: ImageNet~\cite{imagenet_cvpr09}, VOC~\cite{VOC} and COCO~\cite{COCO}.
We found that over 5.3\% of the correct classification results made by the image classification models are not based on object-relevant features. The corresponding rate for the object detection models is over 8.5\%. For specific models, the rate can be as high as 29.1\%.
We additionally defined and demonstrated a simple yet effective strategy to attack deep learning models by leveraging the object relevancy scores.

To summarize, this paper makes three major contributions:
\begin{enumerate}
	\item We proposed a metamorphic testing technique to assess the reliability of inferences generated by deep learning models for image analysis using object-relevant metamorphic relations.
	\item We proposed a metric ``object-relevancy score'' to measure the object relevancy of an inference result. 
	We further show that our metric could be used to effectively facilitate an existing attacking method.
	\item We conducted experiments on 20 common deep learning models for image analysis. We found that the inference results with low object-relevancy scores commonly exist in these models. 
	
\end{enumerate}

\section{Preliminaries}
\label{sec:background}

\subsection{Metamorphic Testing}
Metamorphic testing~\cite{chen1998metamorphic, Chen:2018:MTR:3177787.3143561} was proposed to address the test oracle problem. It works in two steps.
First, it constructs a new set of test inputs (called \textit{follow-up inputs}) from a given set of test inputs (called \textit{source inputs}).
Second, it checks whether the program outputs based on the source inputs and follow-up inputs satisfy some desirable properties, known as \textit{metamorphic relations} (MR).

For example, suppose $p$ is a program implementing the $\sin$ function.
We know that the equation $\sin(\pi + x) = - \sin(x)$ holds for any numeric value $x$.
Leveraging this knowledge, we can apply metamorphic testing to $p$ as follows.
Given a set of source inputs $I_s = \left\{i_1, i_2, \dots, i_n \right\}$,
we first construct a set of follow-up inputs  $I_f = \left\{i'_1, i'_2, \dots, i'_n \right\}$, where $i'_j = \pi + i_j$.
Then, we check whether the metamorphic relation
$\forall j \in \left[1, n\right], p(i_j) = -p(i'_j)$ holds.
A violation of it indicates the presence of faults in $p$. These two steps can be applied repetitively by treating the follow-up inputs in one cycle as the source inputs in the next cycle.
\subsection{Image Analysis Based on Deep Learning}
\label{sec:bg_dl}
Image analysis is a key application of deep learning algorithms to image classification~\cite{resnet, vgg,mobilenet}, object detection~\cite{ssd, fasterrcnn, yolov3}, human keypoint detection~\cite{DBLP:conf/eccv/XiaoWW18, DBLP:conf/cvpr/ChenWPZYS18, DBLP:journals/corr/abs-1904-08900,peng2018megdet} and so on. 
\subsubsection{Image Classification}
An image classifier is built to classify a given image into a category.
AlexNet~\cite{alexnet}, VGG~\cite{vgg}, DenseNet~\cite{densenet}, and sMobileNets~\cite{mobilenet} are popular models for image classification. 
MNIST~\cite{lecun-mnisthandwrittendigit-2010}, CIFAR-10~\cite{cifar10}, and ImageNet~\cite{imagenet_cvpr09} are datasets that have been widely used to evaluate these models.
The performance of the models is mostly evaluated based on the top-1/5 error rate, which refers to the percentage of test images whose correct labels are not in the top-1/5 inference(s) made by models~\cite{alexnet,resnet,vgg,densenet, mobilenet}.

\subsubsection{Object Detection}
An object detector is built to identify the location of target objects in a given image and label their categories.
The object detection result usually contains multiple regions of interest, each of which is marked by a bounding box or a mask, and associated with a confidence value.
Both bounding box and object mask show the region of the object, in the form of rectangle or loop, respectively.
Each region of interest is annotated by a confidence value, indicating the confidence in the inference.
Single Shot MultiBox Detector (SSD)~\cite{ssd}, YOLO~\cite{yolov3} and Faster R-CNN~\cite{fasterrcnn} are popular detectors.
PASCAL VOC~\cite{VOC} and COCO~\cite{COCO} are datasets widely used by studies on object detection. The performance of object detection models is evaluated using several metrics. The VOC challenge uses the metrics \textit{Precision x Recall curve} and \textit{Average Precision}. The COCO challenge uses \textit{mAP} (\underline{m}ean \underline{A}verage \underline{P}recision)~\cite{map}.
The metrics used in both challenges require the computation of IOU (\underline{I}ntersection \underline{O}ver \underline{U}nion):
$ { IOU } = \frac { area\left(b b o x _ { gt } \cap b b o x _ { dt }\right)} { area\left( b b o x _ { gt } \cup b b o x _ { dt }\right) 
}$,
\noindent where the $b b o x _ { gt }$ is the object's bounding box in the ground truth and $b b o x _ { dt }$ is the bounding box of the object detected by a model. 
Here, $b b o x$ can be substituted by object mask $mask$.

\section{object-relevant metamorphic relations}
\label{sec:relation}
With the aim to quantitatively measure to what extend that an inference made by deep learning models is based on object-relevant features, we are motivated to propose two novel metamorphic relations as mentioned in Section \ref{sec:intro}.
This section presents the details of these two relations. 
Specifically, we follow a common metamorphic testing framework to define the metamorphic relations~\cite{Chen:2018:MTR:3177787.3143561}.
In subsequent formulation, let $\mathcal{M}(i)$ denote the inference made by a deep learning model $\mathcal{M}$ on an image $i$, and $\mathcal{D}(\mathcal{M}(i),\ \mathcal{M}(i'))$ denote the distance between two inferences $\mathcal{M}(i)$ and $\mathcal{M}(i')$.

\textbf{MR-1}: 
An image after altering the regions unoccupied by the target object should lead to a similar inference result.

\textit{Relation Formulation}: 
Let $i'_p$ be a follow-up image constructed from a source image $i_p$ for a model $\mathcal{M}$ by preserving the target object but mutating the other parts. We consider such a mutation \textit{object-preserving}.
An example of object-preserving mutation for a keyboard detection model is given by \figurename{\ref{fig:mouse_intro}} (source image) and \figurename{\ref{fig:mouse_intro3}} (follow-up image). 
MR-1 mandates that $\mathcal{M}(i)$ and $\mathcal{M}(i'_p)$ should satisfy the relation: 
$\mathcal{D}(\mathcal{M}(i),\ \mathcal{M}(i'_p)) \leq \Delta_p$. Here, $\Delta_p$ denotes a threshold for the distance between two inference results made by a model under metamorphic testing using object-preserving mutations.

\textit{Explanation}:
If an inference made by a specific model is based on object-relevant features, after object-preserving mutations, the new inference results should be similar since the object-relevant features are preserved and should still be leveraged by the model. 

\textbf{MR-2}: An image after removing the target object should lead to a dissimilar inference result.

\textit{Relation Formulation}: 
Let $i'_r$ be a follow-up image constructed from a source image $i_r$ for a model $\mathcal{M}$ by removing the target object but preserving its background. We consider such a mutation \textit{object-removing}.
An example of object-removing mutation for a keyboard detection model is given by \figurename{\ref{fig:mouse_intro}} (source image) and \figurename{\ref{fig:mouse_intro2}} (follow-up image). 
MR-2 mandates that $\mathcal{M}(i)$ and $\mathcal{M}(i'_r)$ should satisfy the relation: 
$\mathcal{D}(\mathcal{M}(i),\ \mathcal{M}(i'_r)) \geq \Delta_r$. Here, $\Delta_r$ denotes a threshold for the distance between two inference results made by a model under metamorphic testing using object-removing mutations.

\textit{Explanation}:
If an inference made by a specific model is based on object-relevant features, after object-removing mutations, the new inference results should be affected since the object-relevant features disappear and cannot be leveraged by the model. 
\section{Overview}
\label{sec:overview}
\begin{figure}[]
	\centering
	\includegraphics[width=0.48\textwidth]{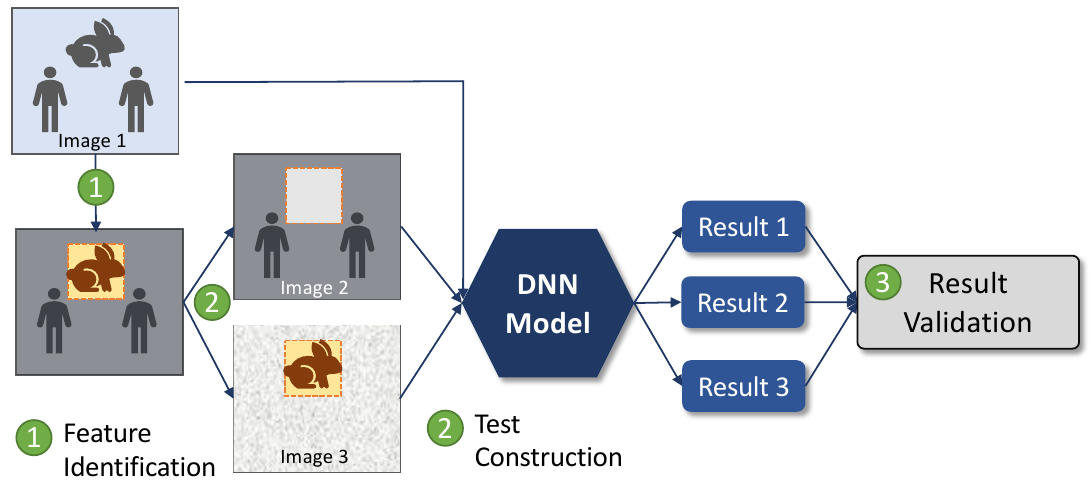}
	\caption{Overview of Our Approach}
	\label{fig:overview}	
\end{figure}
In this section, we present the overview of our approach, which consists of the following three steps:

\textbf{Object-Relevant Feature Identification}:
We treat the images in a given model's validation/test set as source images.
We apply image analysis techniques to each source image and identify its object-relevant features.
Specifically, we leverage the ground truth in the validation/test set to divide an image semantically into two parts, an \textit{object region} and a \textit{background region}.
We consider those segments (i.e., an area of pixels) belonging to the object region as relevant features and the others irrelevant.

\textbf{Follow-up Tests Construction}: Mutation functions are designed to generate follow-up inputs from the source inputs.
Specifically, we design a set of object-preserving mutation functions for MR-1 and and a set of object-removing mutation functions for MR-2. 
The details of these functions are explained in Section~\ref{sec:approach}.

\textbf{Test Result Validation}:
We define distance functions $\mathcal{D}$ for image analysis tasks and object detection tasks, respectively. 
We validate if the distance between the result of a source input and that of its follow-up input fulfills the metamorphic relations as described in Section~\ref{sec:relation}.
Finally, we define the object-relevancy score as a metric to measure to what extent an inference is based on object-relevant features.

\section{Approach}
\label{sec:approach}
We present the details of our approach for two common image analysis tasks in deep learning: image classification and object detection. 
\subsection{Image Classification}
\label{sec:img_cls}
\subsubsection{Object-Relevant Feature Identification}

Since the images used for image classification usually contain one object, we mark the pixels where the object resides as the object region and the remaining pixels as the background region. 
For an image whose ground truth indicates multiple objects, we examine whether any one labels in the ground truth are ranked top-5 by the model.
If so, we regard the object whose label has the highest rank as the object region.
All other objects together with the rest of the image are regarded as the background region.
If not, we regard the union of all objects as the object region and the rest as the background region.
We examine `top-5' since existing evaluations mostly consider the top-5 results as discussed in Section~\ref{sec:bg_dl}.
\subsubsection{Follow-up Tests Construction}
We generate follow-up test input images by semantically mutating a source test input image using the two aforementioned image mutations: \textit{object preserving} mutation and \textit{object removing} mutation.
For each image mutation type, we design multiple mutation functions (e.g., MoveObjToImg), as shown in Table \ref{table:mutation}.
Each of them could use different ingredients (e.g., background image 1).
A mutation function together with an ingredient defines a \textit{mutation operation} (e.g., MoveObjToImg using background image 1).
In total, 38 mutation operations (25 for object preserving and 13 for object removing) are designed.
\begin{table*}[]
	\caption{Image Mutation Functions \& Operations}
	\label{table:mutation}
	\centering
	\scriptsize
	\def\arraystretch{0.7}
	\setlength\tabcolsep{8.8pt}
	\begin{threeparttable}[t]
		
		\begin{tabular}{@{}cccc@{}} 
			\toprule
			\tabincell{c}{Mutation Function \\ Type}                    & \tabincell{c}{Mutation Function \\ Name} & \tabincell{c}{Description} & \tabincell{c}{The Number of \\Mutation Operations}\\ \midrule
			
			& MvObjToImg           & \tabincell{c}{First, directly move the object to a new background image. \\Second, blur the object boundary by median filter.}  & \tabincell{c}{12$^\ddagger$}  
			             
			\\ \cmidrule(l){2-4} 
			\tabincell{c}{Object\\Preserving} 
			& BldObjToImg          & \tabincell{c}{Use OpenCV::seamlessClone to blend the object with a new background image.}  & \tabincell{c}{ 12$^\ddagger$} 
			\\ \cmidrule(l){2-4} 
			& PsvObj            & \tabincell{c}{First, change the value of pixels in the background region to gray. \\Second, blur the object boundary by median filter$^\star$.} & 1
			
			\\ \midrule
			& RmvObjByRGB        & \tabincell{c}{Remove the object by inpainting pixels of the object using a specific color. \\ Then, blur the object boundary by median filter. } & \tabincell{c}{9$^\natural$}
			
			\\ \cmidrule(l){2-4} 
			\tabincell{c}{Object\\Removing} 
			& RmvObjByTool       & \tabincell{c}{Remove the object by inpainting the pixels of the object by the existing tools. \\ We use two tools from OpenCV:  INPAINT\_NS and INPAINT\_TELEA.}  & \tabincell{c}{ 2$^\P$}          
			\\ \cmidrule(l){2-4} 
			& RmvObjByMM       & \tabincell{c}{Remove the object by inpainting the object with the mean/median value of \\ all pixels in the margin between mask and the bounding box. \\This operation is only applicable if both mask and the bounding box exist.} & 2$^\S$          
			\\ \bottomrule	
		\end{tabular}%
		\begin{tablenotes}
			\item[] $\star$:https://en.wikipedia.org/wiki/Median\_filter;  $\S:$  mean and median; $\P:$  2 tools from OpenCV;  $\natural:$  9 common RGB colors.
			\item[] $\ddagger:$ top 12 different images by search "background" and online;
			\setlength{\multicolsep}{0cm}
		\end{tablenotes}
	\end{threeparttable}
\end{table*}

\subsubsection{Metamorphic Relation Validation}
Before formulating the object-relevancy score for an inference, let us introduce our distance function.

\textbf{Distance Function:} 
Given a source input image $i$, an image classification model will generate a probability vector, $m(i) $ = $\left(p_1, p_2, p_3, \ldots\right)$, where $p_k$ denotes the probability that this image belongs to label $ l_{k} $. 
Suppose the image belongs to the label $l$ (i.e., ground truth). Its probability $p$ is the $j$-th largest element in the vector $m(i)$, i.e. rank $j$.
After feeding the follow-up image $i'$ into the model, suppose the new result generated by the model is $m(i')$ = $(p'_1, p'_2, p'_3, \ldots)$.
Similarly, each element $p'_k$ is associated with a specific label $ l_{k} $. 
We assume that in $m(i')$, the ground truth label $l$ has probability $p'$ and its rank is $j'$.
We then compute the distance between $m(i)$ and $m(i')$, according to the type of construction function as follows:

\textit{Object Preserving:} 
If $i'$ is constructed by an \textit{object preserving} function, we follow the convention in Section~\ref{sec:relation} and denote it as $i'_p$. 
We measure the differences using changes of the prediction probability and the rank of the label $l$ as follows:	
{\small
$\mathcal{D}(m(i),\ m(i'_p)) =\left[ 1-\frac{max\left(0, ~ p  -  p ' \right)}{ p } \right] *  \left[ 1 - max\left(0, \frac{1}{j} - \frac{1}{j'}\right) \right] $
}

\noindent The first factor captures on the change in the probability value while the second 
captures the change in the rank.
If this inference is made by a model based on object-relevancy features, there should be no changes in the probability and the rank of $l$, and hence $\mathcal{D}(m(i),\ m(i'_p))$ should be 0.

\textit{Object Removing:} 
We measure how much the prediction probability and the rank of label $ l $ are lowered.
In the ideal case, since the object has been removed, the new probability should be reduced to 0 and $l$'s rank should be lowered. 
	$ \mathcal{D}(m(i),\ m(i'_r)) = \left[\frac{max\left(0, ~ p  - p' \right)}{ p } \right] *  \left[ max\left(0, \frac{1}{j} - \frac{1}{j^{*}}\right) \right]  $

\textbf{Object-relevancy Score:}
We devise a new metric, called \textit{Object-relevancy Score} to measure to what extent an inference $m(i)$ by model $m$ on input $i$ is based on object-relevant features, by integrating the distances between $m(i)$ and $m(i')$ for each follow-up input $i'$. 
 
We define the \textit{Preserving Object-relevancy Score} using the weighted-average of all distances between the source input and each of its follow-up input generated by an \textit{object preserving} mutation operation.
$${ S } _ { p} (m(i))=  \sum w(i'_p) \mathcal{D}(m(i),\ m(i'_p)) $$

Assume $i'_p$ is an image constructed by the $h$-th mutation operation of $n$-th preserving mutation function, its weight $w(i'_p)$ is defined as $w(i'_p) = \frac{1}{H_n*N}$. $H_n$ is the total number of mutation operations in $n$-th mutation function and $N$ is the total number of mutation functions.

Similarly, we define \textit{Removing Object-relevancy Score} as follows:
$${ S } _ { r} (m(i)) =  \sum w(i'_r) \mathcal{D}(m(i),\ m(i'_r)) $$
Again, if $i'_r$ is an image constructed by the $h$-th mutation operation of $n$-th removing mutation function, its weight $w(i'_r)$ is defined as $w(i'_p) = \frac{1}{H_n*N}$. $H_n$ is the total number of mutation operations in $n$-th mutation function and $N$ is the total number of mutation functions.

Finally, we define the \textit{Object-relevancy Score} as follows:
$${ S } (m(i)) = \frac{{ S } _ {p} (m(i)) + { S } _ { r} (m(i))}{2} $$
\subsection{Object Detection}
\label{sec:obj_det}
Object detection task has following differences with image classification problem and thus our the approach need to be changed accordingly.

\textit{Model Output}:
In image classification, model predicts a label for whole image. 
In object detection, given a image, model generates a \textit{result} that contains multiple \textit{record}s.
Each record is a tuple representing a object detected, which contains the detected object's bounding box (or/and mask), label and the corresponding confidence probability.

Therefore, in our approach, when comparing the output between source input and follow-up input, the smallest element for comparison is each record, instead of each result.
We will measure to what extend a record is based on object-relevant features, instead of a result.

Further, we need to effectively map the records from the new results to original results, in order to select the corresponding record to compare with the record to be measured.
Later, we would introduce how we solve this mapping problem by a new concept \textit{Associated Object}. 

\textit{Dataset}:
Compared with image classification, the dataset for object detection has more objects per images.
For example, in the COCO test set, each image contains 7.3 objects on average, shown in Table~\ref{table:dataset}.
In particular, it is common that there exists multiple objects with the same label in a single image.

Such difference brings a new challenge in feature identification.
If we treat all objects as the object region and mutate them together, noises might be introduced when comparing the output of source input and follow-up input. 
For example, assume we want to measure a record `dog' in image to what extend it is based on object-relevant features and there are multiple dogs in this image.
If we simply mutate all `dog's together in this images, this record could be affected.
It is hard for us to understand which dogs' features cause such affection.
For example, it could be the dog having overlap with this record, or it could be another dog which is in the corner of image and far from the region labeled by this record.
We leverage \textit{Associated Object} to solve this challenge, and its detail is followed.

\textbf{Associated Object}
Given a detection record $rcd$, we first locate the object in the ground truth that is most matched to this record, and denote it as the \textit{Associated Object}.

We describe the method to find the \textit{Associated Object} for a given detection record $rcd$ as follows.
Suppose that the input image contains $n$ objects in the ground truth.
For $k$-th object object, we mark it as $ obj_{k} $, with label $ l_k $, bounding box $ bbox_k $.
Suppose the detected object in $rcd$ is $ obj_{rcd} $, the label is $ l_{rcd} $, the bounding box is $ bbox_{rcd} $.

Then we compute the IOU score between the object $ obj_{dt} $ with each object $ obj_{k} $ if they have same label, \ie $ l_{k}  = l_{rcd}$:
$$  { IOU } _ { k } = \frac { area(b b o x _ { k } \cap b b o x _ { rcd }) } { area(b b o x _ { k } \cup b b o x _ { rcd }) } $$
\noindent We then select the object that has the highest IOU with $ obj_{rcd} $ as the \textit{Associated Object}, which is denoted as $ obj_{aso} $,
and the corresponding IOU is denoted as  ${ IOU } _ { aso }$.

In feature identification, we use the region belonging to \textit{Associated Object} as the object region. 
After feeding the mutation into model, we denote the record which has the highest IOU with \textit{Associated Object} as $rcd'$ and we select $rcd'$ to compare with the original record $rcd$.

\subsubsection{Object-Relevant Feature Identification}
After locating the \textit{Associated Object}, we treat the pixels where it locates as the object region and the other area as the background region. 
\subsubsection{Follow-up Tests Construction}
We apply the same mutation function as stated in Table~\ref{table:mutation}.
\subsubsection{Metamorphic Relation Validation}
First, we define the distance function to compare the record $rcd$ in output of a source input with record $rcd'$ in output of follow-up input.
Second, we define the object-relevancy score of a single detection record.
Based on multiple record, we compute the model object-relevance score.

\textbf{Distance Function:}
For record $rcd'$, we denote its object in as $ obj_{rcd}' $ with label $ l_{rcd}' $, bounding box $ bbox_{rcd}'$.
Similarly, we compute the $ IOU_{aso}' $ as follows:
$$ { IOU } _ { aso }' = \frac { area(b b o x _ { aso } \cap b b o x'_ { rcd_p} )} {  area(b b o x _ { aso } \cup b b o x'_ { rcd_p} )}$$
\noindent We then compute the distance between $rcd$ and $rcd'$ based on different mutation function types as follows:

\textit{Object Preserving:} 
If the mutation function is preserving, We denote the record after mutation as $rcd_p'$. 
The IOU with $ obj_{aso} $ is denoted as $IOU_{aso}'$.
We measure the degree of records, including the detection IOUs and labels, that remain the same after mutation. 
$${ \mathcal{D}}(rcd, rcd_p')   = 
\begin{cases}
1 - \frac{max  \left(0,  { IOU } _ { aso } - { IOU } _ { aso } ^ { * } \right)}{IOU_{aso}}, &\text{if } l _ { rcd } = l _ { rcd' } ^ { * } \\
0, & \text{otherwise}
\end{cases}$$

\textit{Object Removing:} 
We measure the degree of results, including the detection IOUs and labels, that alter after mutation. 
	$$ {\mathcal{D}(rcd, rcd_r') }  =\frac { \max (0,  ~ { IOU } _ { aso } -  { IOU } _ { aso } ^ { ' } ) } {  { IOU } _ { aso } }  $$

\textbf{Object-relevancy Score:}
Similar to image classification task, We devise a new metric, called \textit{Object-relevancy Score} to measure to what extend an inference is based on object-relevant features.
Noted that here the inference refers to one detection record.

Firstly, we define the \textit{Preserving Object-relevancy Score} via weighted averaging all distance of $rcd$ and  $rcd'$s constructed by \textit{object preserving} mutation operation.
$${ S } _ { p} (rcd)=  \sum w(rcd'_p) \mathcal{D}(rcd,~rcd'_p) $$
$ w(rcd'_p)$ is the same weight function as in Section~\ref{sec:img_cls}.

Similarly, we define \textit{Removing Object-relevancy Score} as follows:
$${ S } _ { r} (rcd) =  \sum w(rcd'_r) \mathcal{D}(rcd,\ rcd'_r) $$

Finally, we define the \textit{Object-relevancy Score} as follows:
$${ S }  (rcd) = \frac{{ S }_{p}  (rcd) + { S } _ {r} (rcd)}{2} $$

\vspace{3mm}
\section{Experiment Design}

We conducted experiments with the aim to evaluate the effectiveness and usefulness of our proposed approach.
Specifically, we propose the following two research questions:

First, we investigate the performance of the state-of-the-art deep learning models with respect to their object-relevancy scores via answering the following research question:
 
\textbf{Research Question 1:}
Do the state-of-the-art models make inferences based on object-relevancy features?

We answer this question via investigating the following two sub-questions:

\textit{1). Are correct inferences made by existing models problematic if they have low object-relevancy scores?}

To answer this question, we first selected those correct inferences with high probabilities but with low object-relevancy scores.
We then examined whether such inferences are problematic via manual checking with the help of LIME~\cite{lime}, a visualization tool that can explain an inference result made by deep learning models.
If the answer to this research question is ``yes'', which means that even though the inferences made by existing models are correct with respect to their ``labels'', they might still be problematic.
We are then curious towards the distributions of such inferences with high probabilities but low object-relevant scores.
If they are frequently observed in a certain number of image instances, we should then pay more attention to such images when evaluating new models.
Therefore, we are motivated to propose the second sub-question:

\textit{2) How are the distributions of those inferences with high probabilities but low object-relevant scores among different models and different image instances?}

To answer this question, we evaluated 20 models for tasks of image classification and object detection under three large-scale datasets. 
For each model, we calculated the proportion of correct inferences that have high probabilities but  low object-relevancy scores.

Second, we investigate the usefulness of our proposed approach.
Specifically, we investigate whether the object-relevancy score can be leveraged to attack existing state-of-the-art models.
This is motivated by a previous study~\cite{elephant}, which reveals that existing object detection models can be easily attacked via \textit{image transplanting}. 
In their study, moving the object in an image to another with different background could prevent the detector from successfully recognizing it, and thus to attack existing models.
Our proposed object-relevancy score could guide us to perform the operation of \textit{image transplanting} when generating attacking images since it reveals whether an inference is made majorly based on objects or backgrounds.
Therefore, we propose the following research question:

\textbf{Research Question 2:}
Can the object-relevancy score be used to guide the attack of existing the state-of-the-art models?

To answer this question, we designed a new strategy that can effectively generate new images guided by the object-relevancy score.
We then fed these images to the state-of-the-art techniques with the aim to attack them.

\section{Evaluation I: Effectiveness }

To evaluate the effectiveness of our approach, 
we systematically evaluated the performance of existing the state-of-the-art deep learning models in terms of their object-relevant scores.
\subsection{Experimental Setup}
In this experiment, we evaluated 20 models for image classification and object detection on three different datasets quantitatively with respect to their object-relevancy.
Specifically, we selected 10 models for image classification, which are ResNet~\cite{resnet} (including ResNet-50/101/152), MobileNets~\cite{mobilenet}, VGG~\cite{vgg} (including VGG-16 and VGG-19), DenseNet~\cite{densenet} (including DenseNet-121/161), Squeezenet~\cite{squeezenet} and ResNeXt~\cite{ResNext}.
We chose ImageNet to evaluate the performance of these image classification models.
For object detection, the selected models are SSD~\cite{ssd}, YOLOv3~\cite{yolov3} and Faster R-CNN~\cite{fasterrcnn}.
For SSD and YOLOv3, variants using different feature extraction networks (e.g., MobileNets, VGG-16) were also considered in our evaluation.
We evaluated these object detection models under the COCO and VOC datasets.
The information of the selected datasets is listed in Table~\ref{table:dataset}.
All pre-trained models are obtainable from GluonCV~\cite{he2018bag,zhang2019bag}, which is an open-source model zoo providing implementations of common deep learning models.
All of their implementations have reproduced the results as presented in the original publications. 
\begin{table}[t]
	\centering
	\small
	\def\arraystretch{0.8}
	\caption{Datasets Information}
	\label{table:dataset}
	\begin{tabular}{cccc}
& Image Classification & \multicolumn{2}{c}{Object Detection} \\ \midrule
Name	& ImageNet & COCO  & VOC \\ 
Version & 2012 val & 2017 val & 2007 test \\ \midrule
\# images & 50000  & 5000 &  4952 \\
\# categories& 1000   &  80 & 20 \\
\# objects &      -             &        36781           &     14976    \\
\bottomrule  
\end{tabular}
\end{table}
\begin{table}[t]
	\centering
	\caption{Image Classification Models and Their Accuracy }
	\label{table:cls_model}
	\def\arraystretch{0.8}
	\begin{tabular}{cccc}
		\toprule
		 Model & Hashtag & \tabincell{c}{Top-1 \\Accuracy} & \tabincell{c}{Top-5 \\Accuracy}   \\ \midrule
		DenseNet-121 & f27dbf2d & 0.750 	& 0.923 	\\
		DenseNet-161 & b6c8a957 & 0.777	& 0.938 	\\
		MobileNets   & efbb2ca3 & 0.733  	& 0.913	\\
		ResNet-50    & 117a384e & 0.792 	& 0.946		\\
		ResNet-101   & 1b2b825f & 0.805 	& 0.951	 \\
		ResNet-152   & cddbc86f & 0.806 	& 0.953	 \\
		ResNeXt      & 8654ca5d & 0.807	& 0.952	 	\\
		SqueezeNet   & 264ba497 & 0.561	& 0.791	 	\\
		VGG-16       & e660d456 & 0.732 	& 0.913	 \\
		VGG-19       & ad2f660d & 0.741	& 0.914  	\\
		\bottomrule  
	\end{tabular}
\end{table}
\begin{table}[]
	\centering
	\caption{Object Detection Models and Their mAP}
	\label{tab:det_model}
		\def\arraystretch{0.8}
		\begin{tabular}{@{}cccc@{}}
			Dataset               & Model              & Hashtag & mAP    \\ \toprule
\multirow{4}{*}{COCO} & Faster RCNN  &  5b4690fb    & 0.370                     \\
			& SSD(ResNet-50)     & c4835162 & 0.306                   \\
			& YOLOv3(Darknet-53) & 09767802 & 0.370                    \\ \midrule
			& YOLOv3(MobileNets) & 66dbbae6 & 0.280                   \\
			\multirow{6}{*}{VOC} & Faster RCNN  & 447328d8     & 0.783                   \\
			& SSD(MobileNets)    & 37c18076 & 0.754                   \\
			& SSD(ResNet-50)     & 9c8b225a & 0.801                   \\
			& SSD(VGG-16)        & daf8181b & 0.792                   \\
			& YOLOv3(Darknet-53) & f5ece5ce &0.815                   \\ 
			& YOLOv3(MobileNets) & 3b47835a &0.758                  \\ \bottomrule
		\end{tabular}%
\end{table}

\subsection{Results and Findings}
\label{sec:exp}
\subsubsection{Are correct inferences made by existing models problematic if they have low object-relevancy scores?}

To investigate whether correct inferences made by existing models are problematic when they have low object-relevancy scores,
we selected images which are correctly classified by the image classification model ResNet-152 with high probabilities ($\geq 0.5$) but low object-relevancy scores ($\leq 0.5$) for investigation.
Specifically, for each range $\left[0.1*i,0.1*i+0.1\right)$ for integer $i$ from 0 to 4, we selected 20 images whose object-relevancy scores are in this range. 
In total, 100 images were selected.

We then manually investigated to what extent these inference results are problematic. 
Specifically, we presented the original images and the corresponding explanations generated by LIME to 5 senior undergraduate students.
They were then asked to what extent do they think this inference results are problematic in the range of [0\%,100\%].
The value of $0\%$ refers to `Not Problematic' and $100\%$ refers to `Very Problematic'.
The experiment was conducted for each student individually and they were not aware of the corresponding object-relevancy scores.
For each question, they were also asked towards of their confidence on the estimation.
We filtered out the answers with low confidence ($<50\%$) to control the data quality.

\begin{figure}[]
	\centering
	\includegraphics[width=0.45\textwidth]{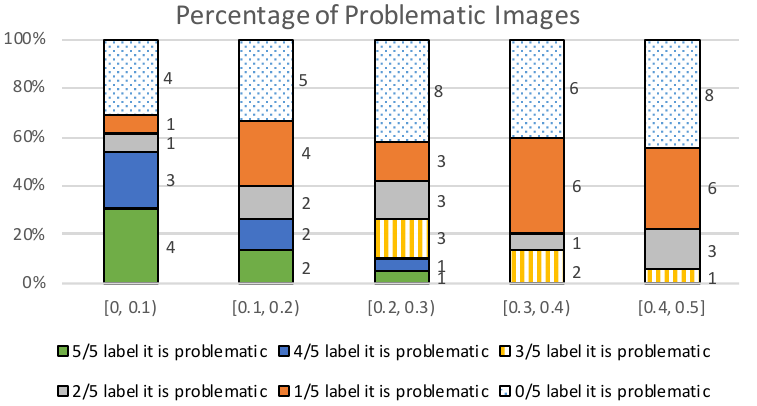}
	\caption{Percentage of Problematic Images}
	\label{fig:manual_check2}
\end{figure}

Finally we collected 80 results and 49 out of them were labeled as problematic ($\geq 50\%$) by at least one student, as shown in~\figurename{\ref{fig:manual_check2}}.
Besides, the lower the object-relevancy score, the more number of students labeled the inferences as problematic. 
For instance, for $53.8\%$ of the inferences whose object-relevancy scores are in the range of $[0, 0.1)$, over 3 students labelled them as problematic. 
Such a ratio is $26.3\%$ for those inferences whose object-relevancy scores are in the range of $[0.2, 0.3)$, and is only $5.6\%$ for the range of $[0.4,0.5)$.

\begin{figure}[]
	\centering
	\begin{subfigure}[b]{0.15\textwidth}
		\includegraphics[width=\textwidth, height=1.9cm]{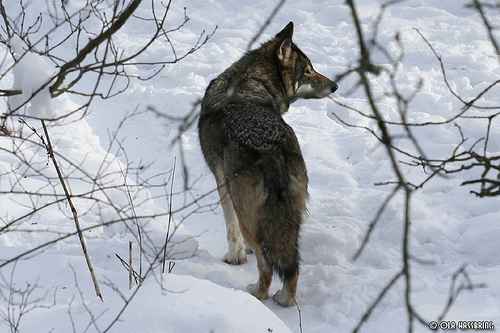}
		\caption{}
	\end{subfigure}
	\begin{subfigure}[b]{0.15\textwidth}
		\includegraphics[width=\textwidth, height=1.9cm]{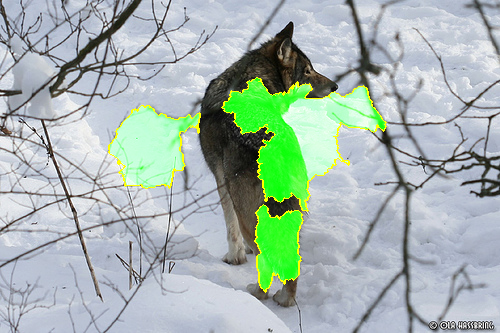}
		\caption{0.553}
	\end{subfigure}
	\begin{subfigure}[b]{0.15\textwidth}
		\includegraphics[width=\textwidth, height=1.9cm]{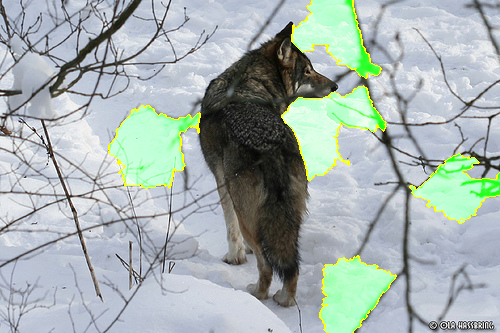}
		\caption{0.399}
	\end{subfigure}
	\begin{subfigure}[b]{0.15\textwidth}
		\includegraphics[width=\textwidth, height=1.9cm]{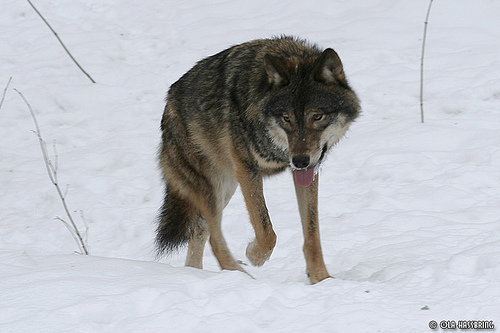}
		\caption{}
	\end{subfigure}
	\begin{subfigure}[b]{0.15\textwidth}
		\includegraphics[width=\textwidth, height=1.9cm]{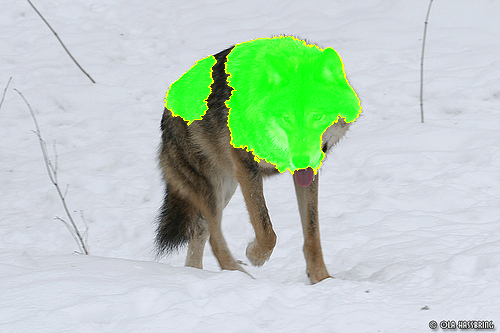}
		\caption{0.671}
	\end{subfigure}
	\begin{subfigure}[b]{0.15\textwidth}
		\includegraphics[width=\textwidth, height=1.9cm]{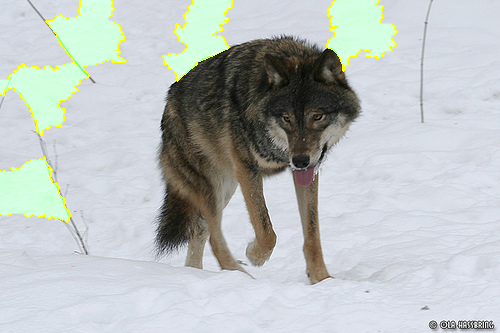}
		\caption{0.451}
	\end{subfigure}
	\begin{subfigure}[b]{0.15\textwidth}
		\includegraphics[width=\textwidth, height=1.9cm]{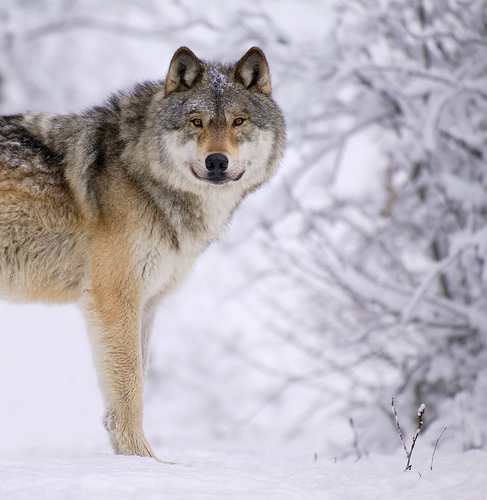}
		\caption{}
	\end{subfigure}
	\begin{subfigure}[b]{0.15\textwidth}
		\includegraphics[width=\textwidth, height=1.9cm]{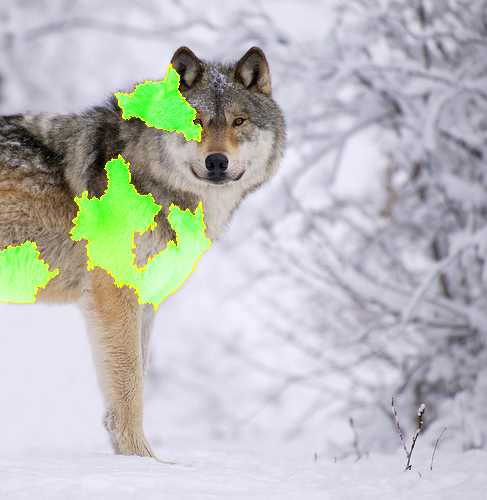}
		\caption{0.703}
	\end{subfigure}
	\begin{subfigure}[b]{0.15\textwidth}
		\includegraphics[width=\textwidth, height=1.9cm]{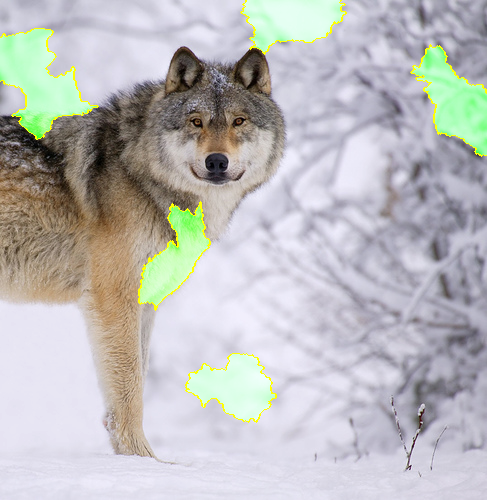}
		\caption{0.472}
	\end{subfigure}
	\caption{Left Column: Test Images. Middle Column: Explanation of Inferences based on Object-Relevant Features. Right Column: Explaination of Inferences based on Object-Irrelevant Features. The Object-Relevancy Scores are Below the Images. The Caption Denotes the Object-Relevancy Score for the Inference.}
	\label{fig:wolf3}
\end{figure}

We also selected three examples to demonstrate that correct inferences with low object-relevancy scores are likely to be problematic as shown in \figurename{\ref{fig:wolf3}}.
The left column in \figurename{\ref{fig:wolf3}} shows the original test images, and the other images show correct inferences that predict the images as ``wolf'' with high probabilities ($\geq 0.5$) made by different models.
The object-relevancy scores are high ($\geq$ 0.5) for the inferences displayed in the middle column while they are low ($\leq$ 0.5) for the inferences displayed in the right column. 
As we can see from the interpretation made by LIME (\ie~the green areas are generated by LIME), 
those correct inferences in the right column are more likely to be problematic since they are majorly made based on object-irrelevant features (\ie~background areas).
Such problematic inferences can also be successfully reflected by their low object-relevant scores.

\subsubsection{How are the distributions of those inferences with high probabilities but low object-relevant scores among different models and different image instances?}
\begin{figure}[]
	\centering
	\includegraphics[width=0.45\textwidth]{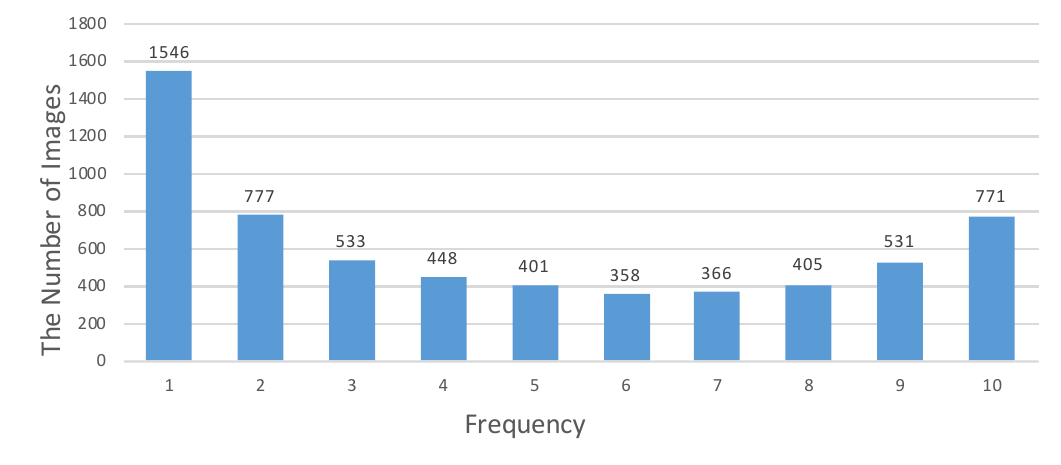}
	\caption{Distribution of Correctly Classified Images with High Probability but Low Object-Relevant Score}
	\label{fig:frequency}
\end{figure}

\begin{figure}[]
	\centering
	\includegraphics[width=0.45\textwidth]{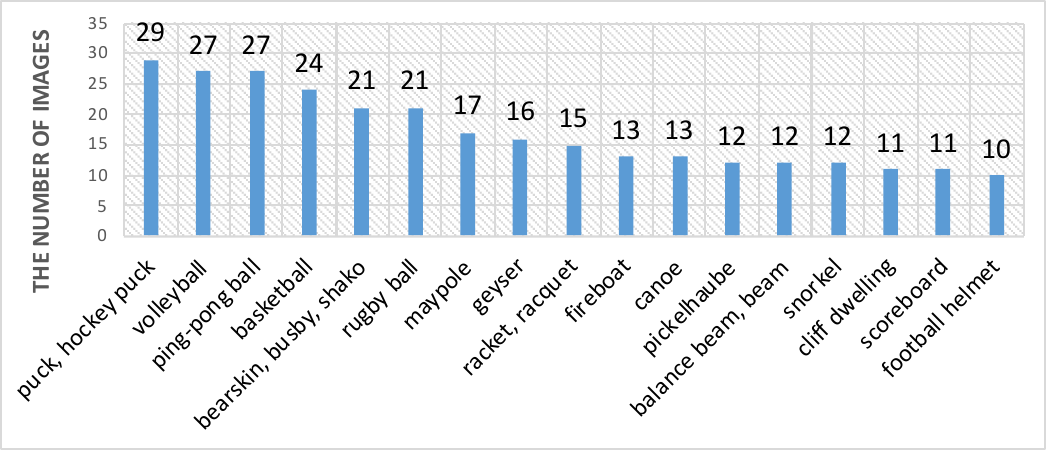}
	\caption{Distribution of Correctly Classified Image with High Probability but Low Object-Relevant Score by all the 10 Models w.r.t Labels (only the frequency larger than 10 are shown).}
	\label{fig:union}
			\vspace{-5mm}
\end{figure}

To investigate the distributions of those inferences that have high probabilities but low object-relevant scores among different models and different image instances, we investigated all the correct (Top-5) inference results with high probabilities ($\geq 0.5$) but low ($\leq 0.5$) object-relevancy scores for each image classification model.
The statistical information of the selected images is displayed in Table~\ref{tab:cls_img_score}.
For each image classification model, around 6\% of their correctly inference results have low object-relevancy scores. 
\begin{table}[]
	\centering
	\caption{The Number and Percentage of Correctly Classified Images with High Classification Probability and Low Inference Object-Relevancy Score in All Correctly Classified Images}
	\def\arraystretch{0.8}
	\label{tab:cls_img_score}
	\begin{tabular}{crrr}
		& Number & Percentage & \tabincell{c}{Total$^\star$}\ \\ \toprule
		DenseNet-121 & 3066 & 6.6\% & 46106 \\
		DenseNet-161 & 2995 & 6.4\% & 46911 \\
		MobileNets   & 2726 & 6.0\% & 45647 \\
		ResNet-50    & 2986 & 6.3\% & 47302 \\
		ResNet-101   & 3171 & 6.7\% & 47551 \\
		ResNet-152   & 3046 & 6.4\% & 47664 \\
		ResNeXt      & 3064 & 6.4\% & 47534 \\
		SqueezeNet   & 2130 & 5.3\% & 40017 \\
		VGG-16       & 2872 & 6.3\% & 45659 \\
		VGG-19       & 2895 & 6.3\% & 45877 \\
		\bottomrule
	\end{tabular}%
	\begin{tablenotes}
		\item[] $\star$: The Number of Correctly Classified Images by This Model
		\setlength{\multicolsep}{0cm}
	\end{tablenotes}
\vspace{-3mm}
\end{table}
We further collected the union of all images with high classification probability but low inference object-relevancy score from 10 models.
In total, we obtained 6317 images and the histogram of these images according to the number of occurrences in the 10 models are shown in~\figurename{~\ref{fig:frequency}}.
It shows that 771 images can be correctly classified with high probabilities ($\geq 0.5$) but the object-relevancy scores evaluated by all the 10 models are low ($\leq 0.5$). 
From the perspective of object-relevancy score, these images should be paid more attentions to since all models evaluated by us do not make inference based on object-relevant features.
We then investigated the distributions of these images, which belongs to 231 distinct labels.
\figurename{\ref{fig:union}} shows the top 17 labels with the highest frequencies in terms of the number of images.
From the perspective of object-relevancy score, these labels should be paid more attention to in future model evaluations.

Similar investigation was conducted for the task of object detection.
We selected all the detection records with high IOU ($\geq 0.5$) but low ($\leq 0.5$) object-relevancy scores.
The statistical information of the selected images is displayed in Table~\ref{tab:det_img_score}.
For most object detection models, around 10\% to 20\% of the correct results have high IOUs but low object-relevancy scores.
\begin{table}[]
	\centering
	\caption{The Number and Percentage of Detected Objects with High IOU and Low Object-Relevancy Score in All Detected Objects with High IOU}
	\setlength\tabcolsep{3.8pt}
		\def\arraystretch{0.8}
	\label{tab:det_img_score}
	\resizebox{0.45\textwidth}{!}{%
		\begin{tabular}{@{}ccrrr@{}}
			Dataset               &      Model              & Number & Percentage & Total$^\star$ \\ \toprule
			\multirow{4}{*}{COCO} & Faster RCNN        & 2648  & 27.0\%     & 9819   \\
			& SSD(ResNet-50)     & 3296  & 21.2\%     & 15557  \\
			& YOLOv3(Darknet-53) & 2248  & 15.1\%     & 14919  \\
			& YOLOv3(MobileNets) & 4450  & 29.1\%     & 15281  \\\midrule
			\multirow{6}{*}{VOC}  & Faster RCNN        & 260   & 28.3\%     & 918    \\
			& SSD(MobileNets)    & 2156  & 21.5\%     & 10046  \\
			& SSD(VGG-16)        & 1162  & 14.4\%     & 8063   \\
			& SSD(ResNet-50)     & 1645  & 16.7\%     & 9821   \\
			& YOLOv3(MobileNets) & 1865  & 17.4\%     & 10708  \\
			& YOLOv3(Darknet-53) & 906   & 8.5\%      & 10610  \\
			\bottomrule
		\end{tabular}%
		
	}
\begin{tablenotes}
	\item[] $\star$: The Number of Detected ($IOU >=0.5$) Objects by This Model
	\setlength{\multicolsep}{0cm}
\end{tablenotes}
\vspace{-5mm}

\end{table}

\section{Evaluation II: Usefulness}
\label{sec:attack}
To demonstrate the usefulness of our proposed approach, 
we designed an approach via leveraging the object-relevancy score to facilitate an existing model attacking approach~\cite{elephant}.
Previous study~\cite{elephant} showed that the state-of-the-art object detection models failed to detect the objects in object-transplanted images, which are generated by replacing an image's sub-regions with another sub-region that contains a object from another image.
The transplanted objects come from the original dataset and they can be correctly detected in their original image.
We extended such attacking method for image classification models.
Specifically, given a model that is trained to classify images of object $l$, there are two attacking scenarios with respect to the two defined metamorphic relations.

\textbf{Scenario 1:}
Synthesize an image having object with label $l$ that force the model incorrectly classify it as other labels.

\textbf{Scenario 2:}
Synthesize an image that does not contain any objects of $l$ but force the model incorrectly classify it as $l$.

In both scenarios, images are synthesized through transplanting as defined in~\cite{elephant}.
More specifically, to realize the first attacking scenario, 
we first select an image $i_a$ from all images with label $l_a$ that is correctly classified by the target model, and another image $i_b$ with another label $l_b$ ($l_a \neq l_b$). 
We then replace the object in $i_b$ with the object extracted from $i_a$ with appropriate adjustment in scale.
We finally feed the synthesized image to the target image classifier.
If the top-1 prediction result is not equal to label $l_a$, we regard it as a successful attack.

\begin{figure}[b!]
	\centering
	\begin{subfigure}[b]{0.15\textwidth}
		\includegraphics[width=\textwidth, height=1.9cm]{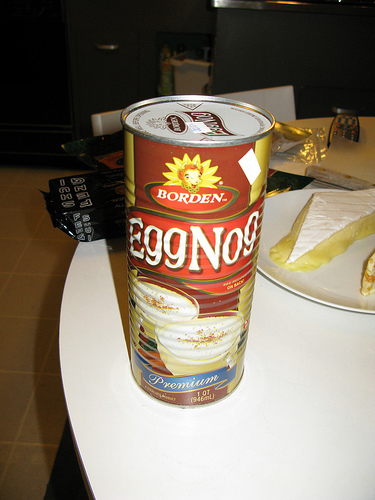}
		\caption{}
		\label{attack_img_a}
	\end{subfigure}
	\begin{subfigure}[b]{0.15\textwidth}
		\includegraphics[width=\textwidth, height=1.9cm]{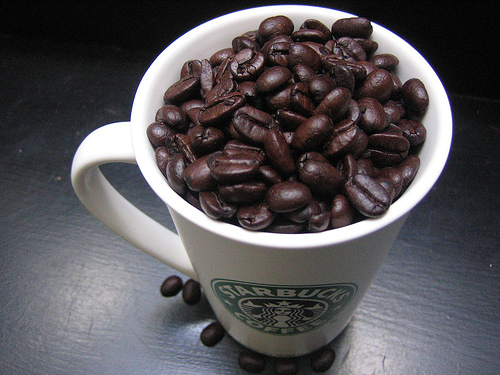}
		\caption{}
		\label{attack_img_b}
	\end{subfigure}
	\begin{subfigure}[b]{0.15\textwidth}
		\includegraphics[width=\textwidth, height=1.9cm]{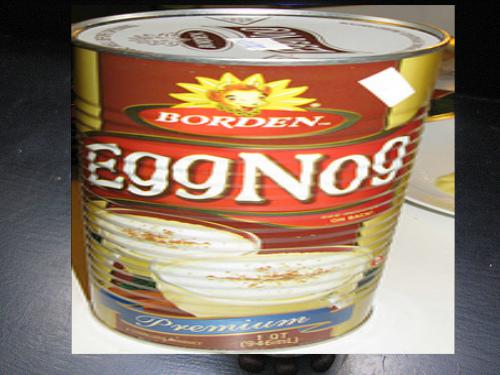}
		\caption{}
		\label{attack_img_ab}
	\end{subfigure}
	\caption{(a): Image $i_a$ with Label `eggnog'. (b): Image $i_b$ with Label `cup'. (c): Successful Attack, which is Predicted as Label `can opener' with a High Probability of 0.9987.}
	\label{attack_img}
\end{figure}

\figurename{\ref{attack_img}} shows an example of such an attacking scenario. 
In this example, we extracted the object in~\figurename{\ref{attack_img_a}} with label `eggnog' and transplanted it to the~\figurename{\ref{attack_img_b}}.
The `cup' in~\figurename{\ref{attack_img_b}} was replaced by `eggnog'.
The synthesized image is shown in~\figurename{\ref{attack_img_ab}} and it is classified as `can opener' by the model.
This is a successful attack, since~\figurename{\ref{attack_img_ab}} has object `eggnog' but it is classified as `can opener' incorrectly instead.

The selection of image $i_a$ and $i_b$ could be either performed randomly or guided by the object-relevancy score.
Intuitively, we should select $i_a$ with a lower preserving object-relevancy score since lower preserving object-relevancy score indicates changing the background of $i_a$ would significantly affect the classification results.
In other word, the object in $i_a$ is ignored by the target model so even a new image has this object, the target model is less likely to recognize it and would not classify it as $l_a$.
Similarly, the $i_b$ should be selected with a lower removing object-relevancy score since it indicates that removing the object from this image would not significantly affect the classification results. 
In other word, the background in $i_b$ affects significantly on the target model's inference.
Therefore, such background could lead the model to label the image as $l_b$, regardless of the real object in the input image.

To guide the synthesis of an attacking image, we sort the images with label $l_a$ according to  their preserving object-relevancy scores, and then select $i_a$ starting from the image with the lowest score.
Similarly, we sort the images with label $l_b$ according to their removing object-relevancy scores, and then select $i_b$ starting from the image with the lowest score. 

\begin{figure}[b!]
	\centering
	\includegraphics[width=0.48\textwidth]{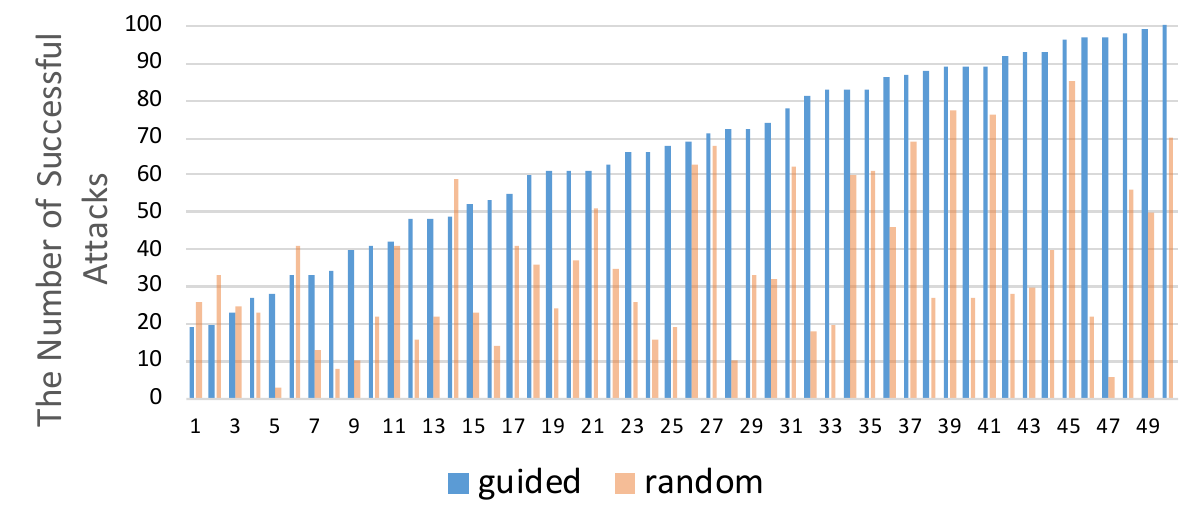}
	\caption{Number of Successful Attacks using Guided Selection and Random Selection for the  Attack Scenario 1}
	\label{fig:attack_1}
\end{figure}

We compared the results using random selection and the guided selection. 
We used the dataset ImageNet and the model ResNet-152 (Model Hashtag: cddbc86f) from GluonCV model zoo~\cite{he2018bag,zhang2019bag}, which achieves the top-5 accuracy on the ImageNet dataset among the whole model zoo (ver0.3.0) in this experimentas.
We randomly generated 50 pairs of labels $l_a$ and $l_b$.
For each pair, we selected two images $i_a$ and $i_b$ (\ie~randomly or guided by the object-relevancy score) 100 times and synthesized 100 images for attacking.
We recorded the number of successful attacks.
The results are displayed in~\figurename{~\ref{fig:attack_1}}.
In total, our guided selection generated 3310 successful attacks (success attack rate $=66.2\%$) while using random selection generated only 1800 successful attacks (success attack rate $=36.0\%$).
For 45 out of the 50 pairs, our strategy is more effective in terms of the number of successful attacks, the improvement of which ranges from 1.02x to 16.17x.
In 40 out of the 45 pairs, our strategy is more efficient since it could synthesize the first successful attacking image more quickly.

\begin{figure}[b!]
	\centering
	\includegraphics[width=0.48\textwidth]{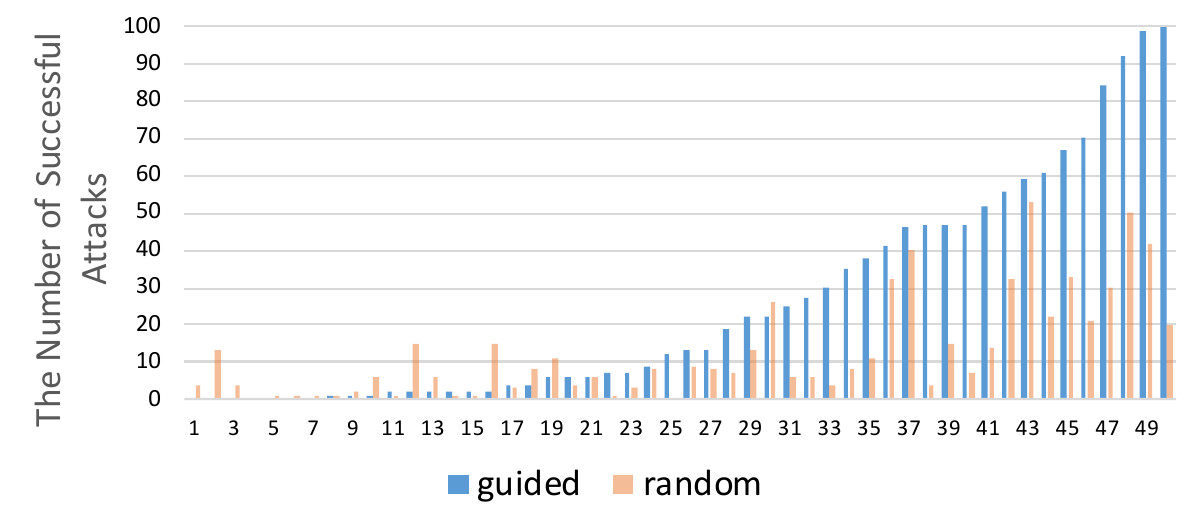}
	\caption{Number of Successful Attacks using Guided Selection and Random Selection for the  Attack Scenario 2}
	\label{fig:attack_2}
\end{figure}

We also designed a similar approach for the second attack. 
First, we selected an image $i_a$ from all images with label $l_a$, and then selected another image $i_b$ with label $l_b$ ($l_a$$\neq$$l_b$). 
After that, we substituted the object in $i_a$ with the object from $i_b$ after appropriate adjustments.
We then fed the synthesized image to an image classifier, and obtained the inference results.
If the top-1 prediction is equal to label $l_a$, we regard it as a successful attack.
Similar to the design of the first attack, in the guided selection, for $i_a$, we prefer those images with lower removing object-relevancy score with respect to the object removing  mutation. 
As for $i_b$, we prefer those images with lower  preserving object-relevancy scores.

We conducted experiments following the methodology of the first attack. The results are displayed in~\figurename{~\ref{fig:attack_2}}.
It shows that our guided selection generated 1288 successful attacks while the random selection only generated 629 successful attacks.
In particular, our guided selection outperformed the random one for 33 out of the 50 pairs.
The two selection strategies achieved the same performance for 3 pairs.
As for the rest 14 pairs, random selection generated 113 successful attacks while our guided selection generated 40 successful attacks.
Although our strategies do not outperform the random selection for certain cases, our guided selection is much more effective in general.

\section{Related Work}
\label{sec:relatedwork}
\subsection{Metamorphic Testing in Deep Learning System}
Several studies have applied metamorphic testing to validate machine learning systems~\cite{issta18meta, XIE2011544,7961649}, including deep learning ones~\cite{issta18meta, Zhang:2018:DGM:3238147.3238187}.
Dwarakanath et al.~\cite{issta18meta} leveraged two sets of metamorphic relations to identify faults in machine learning implementations.
For example, one metamorphic relation for deep learning system is the ``permutation of input channels (i.e. RGB channels) for the training and test data'' would not affect inference results.
To validate whether a specific implementation of DNN satisfies this relation, they re-order the RGB channel of images in both training set and test set. 
They examine the impact on the accuracy or precision of the DNN model after it is trained using the permuted dataset.
Their relations treat the pixels in an image as independent units and they do not consider objects and background in the image.

Xie et al.~\cite{XIE2011544} performed metamorphic testing on two machine learning algorithms: k-Nearest Neighbors and Naïve Bayes Classifier. 
Their work targets at testing attribute-based machine learning models instead of deep learning systems. 
Ding et al.~\cite{7961649} proposed metamorphic relations for deep learning at three different level of validation: system level, data set level and data item level.
For example, a metamorphic relation on system level asserts that DNN should perform better than SVM classifier for image classification.
Both studies require retraining of the machine learning systems under test, they are inapplicable to pre-trained models.

Other studies~\cite{Zhang:2018:DGM:3238147.3238187, deeptest, Zhou:2019:MTD:3314328.3241979} leveraged metamorphic testing in validating autonomous driving systems.
DeepTest~\cite{deeptest} designed a systematic testing approach to detecting the inconsistent behaviors of autonomous driving systems using metamorphic relation.
Their relations focus on general image transformation, including scale, shear, rotation and so on.
Further, DeepRoad~\cite{Zhang:2018:DGM:3238147.3238187} leverage GAN (Generative Adversarial Networks) to improve the quality of transformed image. 
Given a autonomous driving system, DeepRoad mutates the original images to simulate weather conditions such as adding fog to an image. 
An inconsistency is identified if a deep learning system makes inconsistent decision on an image and its mutated one (e.g., the difference of the steering degrees exceeds a certain threshold). 
To the best of our knowledge, we are the first to design metamorphic relations to assess whether an inference is based on object-relevant object or not.

\subsection{Testing Deep Learning Systems}
Besides metamorphic testing, studies have also been made to adapt other classical testing techniques for deep learning systems.
DeepXplore~\cite{deepxplore} proposed neuron coverage to quantify the adequacy of a testing dataset. 
DeepGauge~\cite{Ma:2018:DMT:3238147.3238202} proposed a collection of testing criteria.
DeepFuzz~\cite{odena2018tensorfuzz} and DeepHunt~\cite{DBLP:journals/corr/abs-1809-01266} leveraged fuzz testing to facilateing the debugging processing in DNN.
DeepMutation~\cite{DBLP:conf/issre/MaZSXLJXLLZW18} applied mutation testing to measure the quality of test data in deep learning.

Our study falls into the research direction of testing deep learning systems. 
The major contribution of our study is to test deep learning systems from a new perspective, \ie~the object relevancy of inferences.
This new perspective has not attracted enough attention from communities.
\section{Threats to Validity}
The validity of our study is subject to the following two threats.
First, we collected and tested 20 models for image classification and object detection. These models may not include all models used by deep learning applications. To mitigate the threat, all models collected are representative, designed over popular model architectures in image analysis. We ensured that all models in our implementation achieved an accuracy not worse than the one reported in their original research publications.
All the models collected by us are the representative and popular model architectures in their areas.
Second, our manual check is subject to human mistakes. 
To address the threat, all results are cross validated by 5 senior students independently.
They were not aware of the object-relevancy scores of dataset.

\section{Conclusion}
In this work, we proposed to leverage metamorphic testing to test whether the inference made by pre-trained deep learning models are based on object-relevant features.
We proposed two novel metamorphic relations, from the perspective of object relevancy. 
We devised a metric, \ie~object-relevancy score to measure to what extend an inference is based on object-relevant features.
We applied our approach to 20 popular deep learning models, with 3 large-scale datasets.
We found that the inferences based on object-irrelevant features commonly exist in the output of these models.
We further leveraged the object-relevancy score to facilitate an existing attacking method.

\balance
\bibliography{metaImgModel}{}
\bibliographystyle{IEEEtran}
\end{document}